%% file: main.tex
\documentclass[10pt,twocolumn,letterpaper]{article}

\usepackage[pagenumbers]{cvpr} 

\input{preamble}

\def\paperID{19264}
\def\confName{CVPR}
\def\confYear{2026}

\title{\method: Tracking by Warping instead of Correlation}

\author{
    Zihang Lai$^{1,2}$\qquad Eldar Insafutdinov$^{1}$\qquad Edgar Sucar$^{1}$\qquad Andrea Vedaldi$^{1,2}$\\[1ex]
    {$^{1}$Visual Geometry Group, University of Oxford}\qquad
    {$^{2}$Meta AI}\\[1ex]
}

\begin{document}
\twocolumn[{
\maketitle
\begin{center}
\vspace{-1em}
\includegraphics[width=\textwidth]{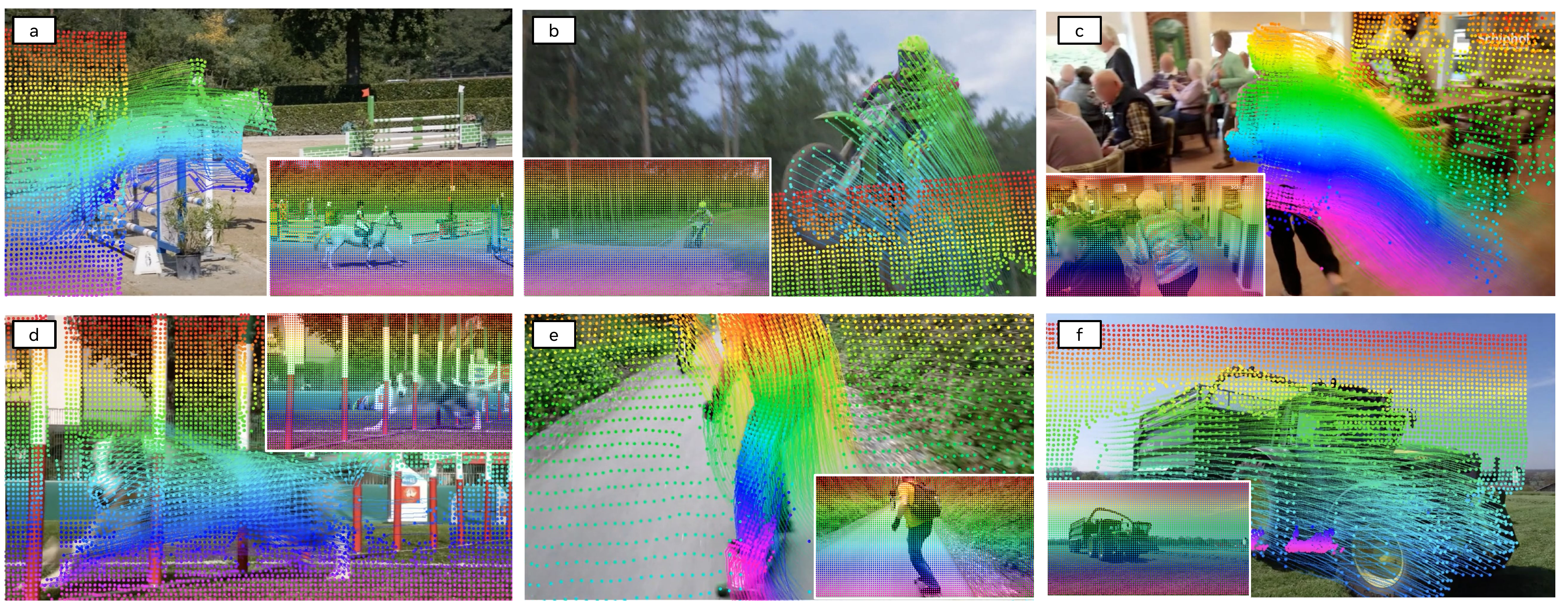}
\vspace{-2em}
\captionof{figure}{\textbf{\method\ results.}
Our tracker produces dense, long-range point tracks in diverse real-world scenes.
It reliably follows humans undergoing rapid motion \textbf{(a,b,c,e)}, animals under repeated occlusions and background clutter \textbf{(d)}, and vehicles in challenging outdoor settings \textbf{(b,f)}.
Corner images show the query points from frame~0. Results are subsampled by a factor of~8 (showing only 1/64 of the predicted points).
}%
\label{fig:splash}
\end{center}
}]
\input{sec/0_abstract}    
\input{sec/1_intro}
\input{sec/2_related}
\input{sec/3_method}
\input{sec/4_experiments}

\input{sec/5_conclusions}

{
    \small
    \bibliographystyle{ieeenat_fullname}
    \bibliography{vedaldi_general,vedaldi_specific,main}
}

\input{sec/X_suppl}

\end{document}

%% file: preamble.tex
\usepackage{booktabs}
\usepackage{caption}
\usepackage{colortbl} 
\usepackage{float}
\usepackage{makecell}
\usepackage{multirow}
\usepackage{pifont}
\usepackage{subcaption}
\usepackage{textpos}
\usepackage{xspace}
\usepackage{placeins}
\usepackage{graphicx}
\usepackage{amsmath}
\usepackage{amssymb}

\usepackage[table]{xcolor}

\definecolor{cvprblue}{rgb}{0.21,0.49,0.74}
\usepackage[pagebackref,breaklinks,colorlinks,allcolors=cvprblue]{hyperref}

\newcommand{\method}{CoWTracker\xspace}

\newcommand{\davgvis}{\ensuremath{\delta_\text{avg}}\xspace}

\newcommand{\ua}{$\uparrow$}
\newcommand{\da}{$\downarrow$}
\newcommand{\warp}{\mathcal{W}}
\newcommand{\z}{z}

\newcolumntype{x}[1]{>{\centering\arraybackslash}p{#1pt}}
\newlength\savewidth
\newcommand{\tablestyle}[2]{\setlength{\tabcolsep}{#1}\renewcommand{\arraystretch}{#2}\centering\footnotesize}

\makeatletter
\renewcommand{\paragraph}{%
  \@startsection{paragraph}{4}%
  {\z@}{-0.5em}{-0.5em}%
  {\normalfont\normalsize\bfseries}%
}
\makeatother

%% file: sec/0_abstract.tex
\begin{abstract}
Dense point tracking is a fundamental problem in computer vision, with applications ranging from video analysis to robotic manipulation. State-of-the-art trackers typically rely on cost volumes to match features across frames, but this approach incurs quadratic complexity in spatial resolution, limiting scalability and efficiency.
In this paper, we propose \method, a novel dense point tracker that eschews cost volumes in favor of warping.
Inspired by recent advances in optical flow, our approach iteratively refines track estimates by warping features from the target frame to the query frame based on the current estimate.
Combined with a transformer architecture that performs joint spatiotemporal reasoning across all tracks, our design establishes long-range correspondences without computing feature correlations.
Our model is simple and achieves state-of-the-art performance on standard dense point tracking benchmarks, including TAP-Vid-DAVIS, TAP-Vid-Kinetics, and Robo-TAP\@.
Remarkably, the model also excels at optical flow, sometimes outperforming specialized methods on the Sintel, KITTI, and Spring benchmarks.
These results suggest that warping-based architectures can unify dense point tracking and optical flow estimation. Project website: \url{cowtracker.github.io}.
\end{abstract}

%% file: sec/1_intro.tex
\section{Introduction}%
\label{sec:intro}

\begin{figure*}[ht]
    \centering
    \includegraphics[width=0.99\linewidth]{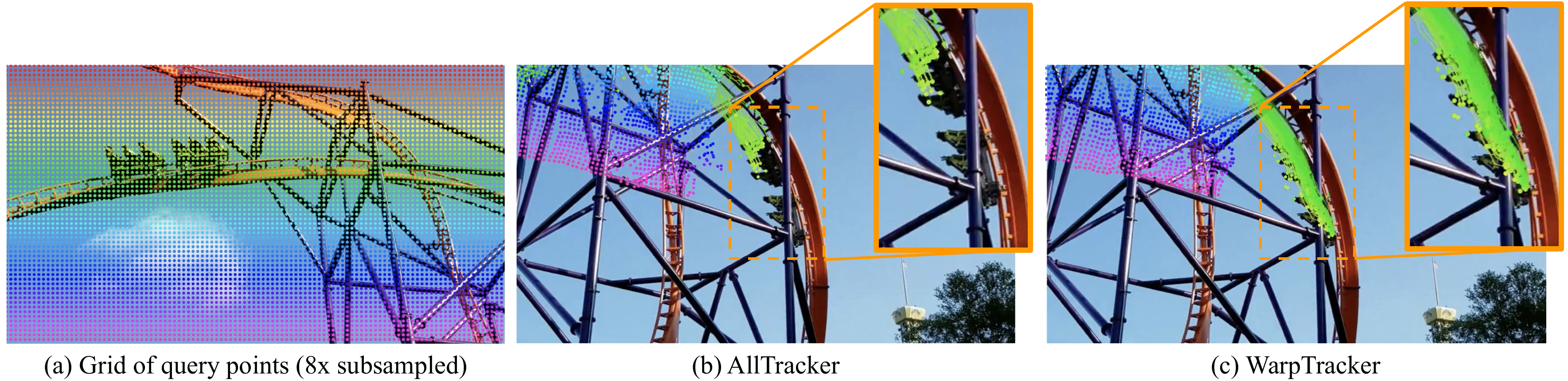}
    \vspace{-0.5em}
    \caption{\textbf{Dense point tracking on a challenging roller-coaster scene.} 
    (a) Initial grid of query points (8$\times$ subsampled). 
    (b) AllTracker~\cite{harley25alltracker:} struggles with the thin strcutures, large viewpoint changes, and occlusions, and fails to track the front half of the coaster (see zoom-in). 
    (c) \method accurately follows the front segment and maintains accurate tracks along the coaster tracks, even near boundaries and through occlusions.}%
    \label{fig:placeholder}
    \vspace{-1.5em}
\end{figure*}

Point tracking~\cite{shi94good} is as old as computer vision and yet remains an important tool in video analysis today.
While early trackers were based on first principles and handcrafted designs, deep learning has transformed the field, starting with the introduction of correlation filters~\cite{bertinetto17end-to-end}.
Most recent progress in point tracking has focused on the \emph{track-any-point} (TAP) formulation of~\cite{harley22particle,doersch22tap-vid:}, where a neural network is tasked with tracking arbitrary points in videos.
The PIPs model introduced by~\cite{harley22particle} has had a strong influence on subsequent work.
A key aspect of its design is the use of a transformer network to reason about tracks over time, treating each point track as a token sequence.
Another key aspect, borrowed from the optical flow literature, is the use of cost volumes~\cite{sun18pwc-net:,dosovitskiy15flownet:,teed20raft:}, which measure similarities between image features across frames and facilitate point matching.

Correlating features and computing cost volumes are staples of optical flow and tracking.
Even so, they can be a bottleneck: matching each feature in one image to each feature in the other scales quadratically with image resolution.
Recently, however, some optical flow works have questioned this design choice.
For example,~\cite{kiefhaber25removing} shows that an optical flow predictor can gradually reduce its dependence on cost volumes during training, discarding them at test time.
More relevant to our work, WAFT~\cite{wang25waft:} shows that cost volumes can be replaced by a warping-based mechanism that is simpler and more efficient.

In this paper, we ask whether these benefits can be transferred from optical flow to dense point tracking.
To answer this question, we propose \method, a novel model for dense point tracking that dispenses with cost volumes and instead uses a warping mechanism similar to WAFT\@.

WAFT borrows from early optical flow approaches such as~\cite{lucas81an-iterative,memin98a-multigrid,brox04high}, and is also related to the Spatial Transformer~\cite{jaderberg15spatial}.\footnote{Not to be confused with the unrelated concept of transformer networks.}
In a nutshell, WAFT iteratively refines the correspondences between two images.
At each iteration, dense features in the second (target) image are warped back to the first (source) image based on the current estimate of the forward flow.
After warping the target features, they are concatenated channel-wise with the corresponding source features and processed by a network to update the flow estimate.
Unlike cost volumes, which explicitly compare each source location to a neighborhood of candidate locations in the target image, this approach evaluates only a single pairing, as specified by the current flow estimate.

At first glance, this design may seem unlikely to establish matches effectively, since each source location is paired with only a single target location at each iteration.
However, another insight from WAFT is that, if the paired features are then processed by a transformer via self-attention, the model can still reason globally about correspondences.

Beyond optical flow, the idea of processing matches globally via self-attention was also proposed in tracking, for example, by CoTracker~\cite{karaev24cotracker}.
Building further on this observation, we draw a connection between joint tracking and warp-based matching.
In fact, joint tracking in CoTracker is obtained by introducing self-attention layers that operate across tracks, which is conceptually equivalent to the self-attention layers used in WAFT to reason globally about correspondences.

Building on this connection, we propose a tracking-by-warping technique.
We estimate tracks densely for every pixel in a reference image.
Based on the current track estimates, we warp features from \emph{all} other frames to the reference frame.
Then, we use a transformer with separate self-attention along the spatial and temporal dimensions to refine the tracks.
This is inspired by the design of PIPs and CoTracker, and it also subsumes the spatial attention in WAFT\@.

The resulting design is extremely simple yet effective.
When combined with powerful feature encoders---especially VGGT~\cite{wang25vggt}---\method achieves state-of-the-art performance on several dense point tracking benchmarks, including
TAP-Vid-DAVIS~\cite{doersch22tap-vid:},
TAP-Vid-Kinetics~\cite{harley22particle}, and
Robo-TAP~\cite{vecerik2024robotap}.
Moreover, the \emph{same model} also performs strongly on optical flow, sometimes outperforming the state of the art on benchmarks such as
Sintel~\cite{butler12a-naturalistic},
KITTI~\cite{geiger12are-we-ready}, and
Spring~\cite{mehl23spring:}.

In summary, our contributions are:
(i) we show that warp-based matching works well for optical flow and tracking, and leverage this observation to build \method, a unified model that solves both while eschewing cost volume computation;
(ii) we demonstrate that \method achieves state-of-the-art performance on several dense point tracking benchmarks; and
(iii) we show that the \emph{same} model is also highly competitive for optical flow, sometimes outperforming the state of the art on standard benchmarks for this task.
We believe this design can serve as a stepping stone for future research in optical flow, point tracking, and other matching tasks.

%% file: sec/2_related.tex
\section{Related work}%
\label{sec:related}

While 2D tracking has a long history in computer vision~\cite{shi94good}, we focus on the so-called \emph{tracking any point} (TAP) problem, popularized by recent works~\citep{harley22particle,doersch22tap-vid:} and most relevant to our setting.
The former also revisits Particle Video~\citep{sand08particle} using deep learning; the resulting PIPs tracker introduces several design elements that influenced numerous follow-ups and extensions.
PIPs, inspired by the optical flow method RAFT~\citep{teed20raft:}, computes correlation maps between image features (a \emph{correlation volume}) and refines track estimates using a transformer network.

TAP-Vid~\citep{doersch22tap-vid:} refines the TAP problem statement, proposes multiple benchmarks, and introduces TAP-Net, a lightweight model for point tracking.
TAPIR~\citep{doersch23tapir:} fuses TAP-Net-style global matching with PIPs-style local refinement, yielding substantial accuracy gains.
Zheng et al.~\citep{zheng23pointodyssey:} contribute the PointOdyssey synthetic benchmark and present PIPs++, an improved PIPs variant for longer-term tracking.

CoTracker~\citep{karaev24cotracker,karaev25cotracker3} exploits correlations between multiple simultaneous tracks to improve robustness under occlusion and out-of-frame motion; CoTracker3 simplifies the architecture and proposes a data-efficient self-training regime.
DOT~\citep{le-moing24dense} builds on CoTracker by densifying its outputs, while VGGSfM~\citep{wang24vggsfm:} adopts a coarse-to-fine design that validates tracks via 3D reconstruction (but targets static scenes only).
BootsTAPIR~\citep{doersch24bootstap:} further improves TAPIR through large-scale self-training on millions of videos.

Inspired by DETR~\citep{carion20end-to-end}, TAPTR~\citep{li24taptr:} formulates point tracking as an end-to-end transformer, where points are decoder queries.
TAPTRv2~\cite{li2024taptrv2} introduces a deformable-attention mechanism that eliminates cost volumes for sparse point tracking, but is difficult to extend to dense tracking because each query outputs only a single displacement.
LocoTrack~\citep{cho24local} extends 2D correlation features to 4D correlation volumes, simplifying the pipeline and improving efficiency.
TAPVid-3D~\citep{koppula24tapvid-3d:} is a 3D-aware variant of TAP-Vid that incorporates reconstruction or multi-view cues.
TAPIP3D~\citep{zhang25tapip3d:} further extends TAP methods to 3D/depth-aware tracking.
TAPNext~\citep{zholus25tapnext:} focuses on improved matching and temporal modeling for higher accuracy and speed.
ReTracker~\citep{tan2025retracker} boosts accuracy via pretraining on large-scale image-matching datasets.
DELTA~\citep{ngo24delta:} develops dense embedding and association learning for long-term robustness.
SceneTracker~\citep{wang24scenetracker:} leverages scene-level geometry to validate and recover tracks,
and AllTracker~\citep{harley25alltracker:} proposes a unified model for tracking all points or objects in a video.
In this work, we depart from the cost-volume-based paradigm adopted by most prior TAP approaches, opting for a purely warping-based tracker that is simple, efficient, and accurate.

%% file: sec/3_method.tex
\section{Method}%
\label{sec:method}

\newcommand{\real}{\mathbb{R}}
\newcommand{\warpedFeatures}{G}

In this section, we introduce \method, a new warping-based dense tracker.
The full model, shown in \cref{fig:update_operator}, consists of three parts:
(1) a backbone that produces low-resolution features for each frame,
(2) a DPT upsampler that lifts those features to the target (higher) resolution for tracking, and
(3) a tracker that computes dense tracks by warping.

\input{figs/fig_update_operator}

\subsection{Problem Definition}%
\label{sec:problem-definition}

Let a video be $\{I_t\}_{t=0}^T$, consisting of a sequence of $T + 1$ images
$
I_t\in\real^{3\times H\times W}
$,
$
t \in \mathcal{T} = \{0, 1, \dots, T \}
$.
$I_0$ is the \emph{query frame} and $I_1,\dots,I_T$ are the \emph{target frames}.
Let $p \in \real^2$ denote a 2D pixel location in image coordinates and let $\mathcal{P} \subset \real^2$ be a subset of locations that we use as \emph{queries}.
The goal of the tracker is to determine the location
$
x_t(p) \in \real^2
$
of each query pixel $p \in \mathcal{P}$ in each target frame $I_t$, $t=1,\dots,T$, under the assumption that the track starts at the query locations at time $t=0$, i.e., $x_0(p) = p$.

For convenience, we express tracks in terms of the displacement field
$
u : \mathcal{T} \times \mathcal{P} \to \real^2
$,
defined by the equation
\begin{equation}\label{sec:displacement-field}
x_t(p) = p + u_t(p),
\quad p \in \mathcal{P},
\quad t \in \mathcal{T}.
\end{equation}
We additionally predict a visibility probability $v_t(p) \in [0,1]$ indicating whether the point is observable in $I_t$, and a confidence score $\tau_t(p) \in [0,1]$.

\subsection{\method Overview}%
\label{sec:overview}

Unlike previous trackers in the vein of PIPs and follow-up works, our architecture eschews the calculation of cost volumes, utilizing a warp-based formulation instead.

The tracker takes the input video $I$ and extracts dense features $F$ using an off-the-shelf feature extractor and an upsampling layer to obtain a spatial resolution sufficient for accurate tracking.
The tracker $\Phi$ then starts by assuming a null displacement field $u = \mathbf{0}$ (corresponding to assuming that points are stationary) and iteratively updates it, captured by the recurrence
$
u' = \Phi(u \mid F)
$.
Key to the tracker is a warping operation that, given the current estimate of the displacement field $u$, samples features from each target frame at locations specified by $u$ and brings them into correspondence with the query frame.
This results in a \emph{warped feature field} $\warpedFeatures = \warp(F, u, p)$, from which the updated displacement field is computed.
The latter uses a transformer network which, crucially, alternates spatial and temporal attention layers.
As noted in CoTracker, tracks are highly correlated.
This correlation is crucial to compensating for the fact that there is no cost volume to explicitly search for matches~\cite{wang25waft:}.

See the Appendix for pseudocode describing this process.
Next, we explain each step in detail.

\subsection{Features}%
\label{sec:features}

\paragraph{Feature Extractor.}

To compute dense features from the input video $I$, we use a network $\Psi$ for dense feature extraction.
In practice, we consider strong pretrained models like VGGT\@.
Because the feature extractor can process all frames jointly (e.g., VGGT), we write this as a map $F=\Psi(I)$ assigning the tensor
$
I \in \real^{T\times 3\times H\times W}
$
of all video frames to a tensor
$
F \in \real^{T\times H'\times W'\times C}
$
of corresponding features with (typically) lower spatial resolution $H'\times W'$ and $C_b$ channels.
The resolution is determined by the stride or patch size $s$ of the feature extractor (typically $14$ or $16$), so that $\hat H = \lfloor H/s \rfloor$ and $\hat W = \lfloor W/s \rfloor$.

Unless otherwise noted, we freeze early stages and optionally fine-tune the last blocks, and do not otherwise change the architecture.

\paragraph{Upsampler.}

Methods that use cost volumes typically operate at low spatial resolution to keep memory consumption manageable.
In contrast, our warping-only design can efficiently handle higher spatial resolution, as it does not require computing and storing large cost volumes.
We therefore upsample the features $\hat F$ to a higher spatial resolution $H' \times W'$ using a DPT upsampler~\cite{ranftl21vision}, resulting in features $F = \operatorname{DPT}(\Psi(I))$ with stride $s' = 2 \ll s$.
This uses skip connections from the backbone and a lightweight convolutional decoder.
Following WAFT, we also use a small U-Net that directly takes raw images as input and concatenates its output with the upsampled backbone features.

Bringing the resolution of the features close to the input video resolution improves tracking of thin structures and near boundaries.
No additional correlation tensor or pyramid is constructed.

\input{tables/tab_tapvid}

\subsection{Warping-only Tracker}%
\label{sec:head}

The high-resolution feature maps produced by the backbone and DPT upsampler are consumed by our \emph{warping-only} tracker to compute dense tracks from the query frame to every target frame.
\Cref{fig:update_operator} depicts one iteration of the update operator inside the tracker head.

The tracker takes as input the feature tensor $F$ computed in \cref{sec:features}.
It also maintains two states.
The first is the current estimate of the displacement fields
$
u^{(k)} : \mathcal{T} \times \mathcal{P} \to \real^{2},
$
i.e., a $|\mathcal{T}|\times|\mathcal{P}|\times 2$ tensor.
The second is a \emph{hidden state}
$
h^{(k)} : \mathcal{T} \times \mathcal{P} \to \real^{D_h},
$
where $D_h$ is the hidden dimension.

The index
$
k \in \{ 0, 1, \dots, K \}
$
denotes the update iteration.
As noted above, we initialize the displacement field by setting it to zero, i.e., $u^{(0)} = 0$.
The hidden state $h^{(0)}$ is initialized from the features $F$ by setting $h^{(0)}_t = \phi(F_0 \oplus F_t)$, where $F_0 \oplus F_t$ is the channel-wise concatenation of the features for frames $0$ and $t$, and
$
\phi: \real^{2C} \rightarrow \real^{D_h}
$
is a small network implemented as a $1\times1$ convolution followed by layer normalization to reduce the number of channels from $2C$ to $D_h$.

The model incrementally \emph{updates} $u$ and $h$ via $K$ iterations of a learned update operator (\cref{fig:update_operator}):
$$
(u^{(k+1)}, h^{(k+1)}) = \Phi(u^{(k)}, h^{(k)} \mid F),
\quad k = 0, \dots, K - 1.
$$
After $K$ iterations, the final displacement field $u = u^{(K)}$ is obtained.
Track visibility and confidence are output by readout heads on the final hidden state $h^{(K)}$ via linear layers followed by a sigmoid:
$$
v_t = \sigma \! \left(h^{(K)} W_v\right),
\quad
\tau_t = \sigma \! \left(h^{(K)} W_\tau\right).
$$

Next, we describe the design of the update operator $\Phi$.
This begins by \emph{warping} the features $F$ based on the current displacement estimate
$
\warpedFeatures = \mathcal{W}(F, u, p),
$
given by
\begin{equation}\label{eq:warp}
\warpedFeatures_t(p) = \operatorname{sample}(F_t,\, p + u_t(p)).
\end{equation}
Here, $\operatorname{sample}(\cdot)$ denotes bilinear sampling.
This aligns the feature vectors in each target frame $t$ to the query frame at location $p$ at time $0$.
Thus, $\warpedFeatures$ is a $|\mathcal{T}|\times|\mathcal{P}|\times C$ tensor of features aligned across time.

The $k$-th update takes the
warped features ${\warpedFeatures}^{(k)} = \mathcal{W}(F, u^{(k)}, p)$,
the query features $F_0$,
the current displacement $u^{(k)}$ and hidden state $h^{(k)}$,
and concatenates them into a tensor $\z \in \real^{|\mathcal{T}|\times H' \times W' \times (2C + D_h + 2)}$ given by:
$$
\z_t
=
{\warpedFeatures_t}^{(k)} \oplus
F_0 \oplus
u^{(k)}_t \oplus
h^{(k)}_t,
\quad
t\in\mathcal{T}.
$$
The features $\z$ are interpreted as a $|\mathcal{T}|\times|\mathcal{P}|$ matrix of tokens organized by time and space.
We augment the features with spatial and temporal embeddings and process these tokens using a Vision Transformer (ViT)~\cite{dosovitskiy21an-image} adapted to video~\cite{arnab21vivit:}, where spatial and temporal attentions interleave: every two spatial self-attention blocks, we insert one temporal attention block.
Each spatial self-attention layer runs independently over $\mathcal{P}$ for each fixed temporal slice $t$, and each temporal attention layer runs over $\mathcal{T}$ for each fixed spatial position $p$.
The architecture otherwise follows ViT, with MLP blocks, residual connections, and LayerNorm throughout.
Finally, a linear head predicts the residual displacement:
\begin{equation}
\label{eq:delta-u}
\Delta u^{(k+1)} = h^{(k+1)} W_u,
\quad
u^{(k+1)} = u^{(k)} + \Delta u^{(k+1)}.
\end{equation}

Note that there is no correlation volume computation in the head; instead, the only place where cross-frame features are paired is the warping operator in \cref{eq:warp}.
Hence, the cost of running the head scales linearly with the number of targets $T$, the feature resolution $|\mathcal{P}|$, and the number of iterations $K$, making it possible to use a relatively large spatial resolution.

%% file: figs/fig_update_operator.tex
\begin{figure*}[t]
\centering
\includegraphics[width=0.99\linewidth]{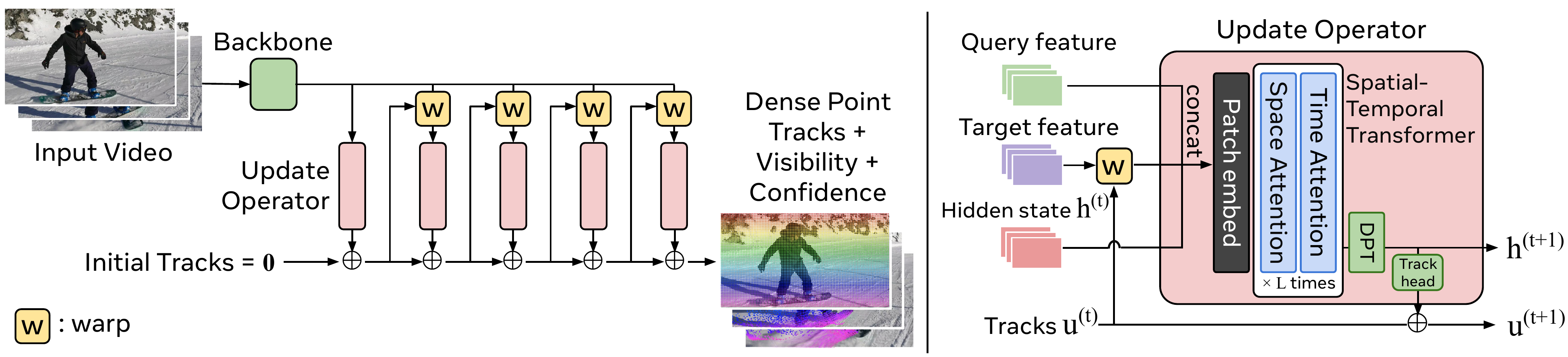}
\vspace{-1em}
\caption{\textbf{Left: \method Pipeline.}
The backbone extracts video features from the input video, and a lightweight update operator (see right for details) iteratively warps and refines tracks to yield dense trajectories, visibility, and confidence.
\textbf{Right: Update operator.}
Warped query/target features, hidden states, and current track estimates are fused by a spatial-temporal transformer to predict residual motion and update hidden states.
}
\vspace{-1em}
\label{fig:update_operator}
\end{figure*}

%% file: tables/tab_tapvid.tex
\begin{table*}[!ht]
\centering
\setlength{\tabcolsep}{4.3pt}
\footnotesize
\begin{tabular}{llcccccccccccccccc}
\toprule
\multirow{2}{*}[-0.2em]{Method} & \multirow{2}{*}[-0.2em]{Train  data}
& \multicolumn{3}{c}{DAVIS}
& \multicolumn{3}{c}{RGB-S}
& \multicolumn{3}{c}{RoboTAP}
& \multicolumn{3}{c}{Kinetics}
& \multicolumn{3}{c}{Mean} \\
\cmidrule(lr){3-5}
\cmidrule(lr){6-8}
\cmidrule(lr){9-11}
\cmidrule(lr){12-14}
\cmidrule(){15-17}
& & AJ\ua & \davgvis\ua & OA\ua
  & AJ\ua & \davgvis\ua & OA\ua
  & AJ\ua & \davgvis\ua & OA\ua
  & AJ\ua & \davgvis\ua & OA\ua
  & AJ\ua & \davgvis\ua & OA\ua \\
\midrule
\multicolumn{17}{l}{\textit{Sparse Trackers}} \\
\midrule
PIPs++~\cite{zheng23pointodyssey:} & PO
& ---  & 73.7 & ---
& ---  & 58.5 & ---
& ---  & 63.5 & ---
& ---  & 63.5 & ---
& --- & 64.8 & --- \\
TAPIR~\cite{doersch23tapir:} & Kub
& 56.2 & 70.0 & 86.5
& 55.5 & 69.7 & 88.0
& 49.6 & 64.2 & 85.0
& 49.6 & 64.2 & 85.0
& 52.7 & 67.0 & 86.1 \\
CoTracker~\cite{karaev24cotracker} & Kub
& 61.8 & 76.1 & 88.3
& 67.4 & 78.9 & 85.2
& 49.6 & 64.3 & 83.3
& 49.6 & 64.3 & 83.3
& 57.1 & 70.9 & 85.0 \\
TAPTR~\cite{li24taptr:} & Kub
& 63.0 & 76.1 & {91.1}
& 60.8 & 76.2 & 87.0
& 49.0 & 64.4 & 85.2
& 49.0 & 64.4 & 85.2
& 55.5 & 70.3 & 87.1 \\
LocoTrack~\cite{cho24local} & Kub
& 62.9 & 75.3 & 87.2
& 69.7 & 83.2 & 89.5
& 52.9 & 66.8 & 85.3
& 52.9 & 66.8 & 85.3
& 59.6 & 73.0 & 86.8 \\
BootsTAPIR~\cite{doersch24bootstap:} & Kub+15\textbf{M}
& 61.4 & 73.6 & 88.7
& 70.8 & 83.0 & 89.9
& 54.6 & {68.4} & 86.5
& 54.6 & 68.4 & 86.5
& 60.4 & 73.4 & 87.9 \\
CoTracker3~(onl.)~\cite{karaev25cotracker3} & Kub+15k
& 63.8 & 76.3 & 90.2
& 71.7 & 83.6 & {91.1}
& {55.8} & {68.5} & {88.3}
& {55.8} & {68.5} & {88.3}
& 61.8 & {74.2} & 89.5 \\
CoTracker3~(offl.)~\cite{karaev25cotracker3} & Kub+15k
& {64.4} & {76.9} & {91.2}
& {74.3} & {85.2} & {92.4}
& {54.7} & 67.8 & {87.4}
& {54.7} & 67.8 & {87.4}
& 62.0 & {74.4} & 89.6 \\
\midrule
\multicolumn{17}{l}{\textit{Dense Trackers}} \\
\midrule
DOT~\cite{moing23dense} & Kub
& 60.1 & 74.5 & 89.0
& 77.1 & 87.7 & 93.3
& --- & --- & ---
& 48.4 & 63.8 & 85.2
& --- & --- & --- \\
DELTA~\cite{ngo24delta:} & Kub
& 60.8 & 75.2 & 87.6
& 72.2 & 83.0 & 91.4
& 60.2 & 73.4 & 85.6
& 51.0 & 66.6 & 84.0
& 61.1 & 74.6 & 87.2 \\
AllTracker~\cite{harley25alltracker:} & Kub
& 61.9 & 75.4 & 87.8
& 80.7 & 90.1 & 91.7
& 70.1 & 81.5 & 92.2
& 59.3 & 71.3 & 89.3
& 68.0 & 79.6 & 90.3 \\
AllTracker~\cite{harley25alltracker:} & Kub+Mix
& 63.3 & 76.3 & 90.1
& 81.1 & 90.0 & 92.8
& 71.9 & \textbf{83.4} & 92.8
& 59.1 & 72.3 & 90.3
& 68.9 & 80.5 & 91.5 \\
\method & Kub
& \textbf{65.5} & \textbf{78.0} & \textbf{92.1}
& \textbf{85.4} & \textbf{92.8} & \textbf{94.9}
& \textbf{73.2} & \textbf{83.4} & \textbf{94.7}
& \textbf{60.9} & \textbf{73.1} & \textbf{91.5}
& \textbf{71.3} & \textbf{81.8} & \textbf{93.3} \\
\bottomrule
\end{tabular}
\vspace{-1em}
\caption{
\textbf{TAP-Vid~\cite{doersch22tap-vid:} (DAVIS, RGB-Stacking, Kinetics) and RoboTAP~\cite{vecerik2024robotap} benchmarks.}
We report Average Jaccard (AJ), \davgvis, and Occlusion Accuracy (OA; \ua higher is better). Trained only on Kubric data, our dense \method achieves the best mean performance across all four datasets and all three metrics, outperforming prior dense methods such as AllTracker as well as strong sparse TAP baselines.
}
\vspace{-1em}
\label{tab:tab_tapvid}
\end{table*}

%% file: sec/4_experiments.tex
\section{Experiments}%
\label{sec:experiments}

\FloatBarrier{}

\input{figs/bmx_comparison}                %

We show that \method demonstrates competitive performance on point tracking (\cref{sec:exp-pt}) and optical flow (\cref{sec:exp-flow}), and we ablate our design choices (\cref{sec:ablation}).

\subsection{Implementation details}%
\label{sec:exp-setup}

\paragraph{Architecture.}

Unless noted otherwise, we use a pretrained VGGT~\cite{wang25vggt} as our backbone and a DPT-style~\cite{ranftl21vision} upsampler to lift low-resolution features to high resolution.
We freeze the patch-embedding layers in VGGT and finetune the other parts of the backbone, together with the upsampler (initialized from scratch) and the tracker.
The tracker uses $K=5$ iterations and operates on features with stride $s'=2$.

\paragraph{Training.}

We train \method on Kubric data with Huber losses for both visible and occluded tracks, with exponentially increasing weights for each iteration, following~\cite{karaev25cotracker3}.
Confidence and visibility are supervised using a binary cross-entropy (BCE) loss at each iterative update.
The ground-truth confidence is given by an indicator function that checks whether the predicted track falls within 12 pixels of the ground-truth track.
We use the AdamW optimizer with learning rate $5\times 10^{-4}$ and cosine decay, a batch size of 32 videos with up to 16 frames of size $336\times 560$, and train for 50k iterations.
We use random frame rate and random video length for data augmentation.
Mixed precision and gradient checkpointing are enabled.

\subsection{Point Tracking}%
\label{sec:exp-pt}

We evaluate the point-tracking performance of \method on the TAP-Vid benchmark suite~\cite{doersch22tap-vid:}---which includes the Kinetics, DAVIS, and RGB-Stacking subsets---and on RoboTAP~\cite{vecerik2024robotap}.
TAP-Vid Kinetics contains 1{,}144 videos from the Kinetics validation set~\cite{carreira2019short} featuring complex camera motion; the DAVIS subset provides 30 real-world videos from DAVIS~\cite{Perazzi2016}; RGB-Stacking consists of synthetic robotic sequences with large texture-less regions; and RoboTAP includes 265 real-world robotic manipulation videos.
We report the standard metrics for TAP-Vid: Occlusion Accuracy (OA), $\davgvis$, and Average Jaccard (AJ).

\subsubsection{Quantitative Results}

\Cref{tab:tab_tapvid} reports results on the TAP-Vid benchmark and RoboTAP\@.
Our model surpasses the strongest dense prior, AllTracker, on all four datasets and across all three metrics.
Averaged over the datasets, \method improves AJ by {2.4}, $\davgvis$ by {1.3}, and OA by {1.8}.
Per dataset, the gains are also consistent.
These improvements are achieved without cost volumes or multi-resolution pyramids (used in AllTracker), indicating that a warping-only head with a lightweight video-based transformer can enable accurate point tracking.

This gap widens further when training data is controlled: against AllTracker (Kub), the mean improvements of \method increase to 3.3 / 2.2 / 3.0 in AJ/$\davgvis$/OA, respectively.
We attribute robustness across heterogeneous domains---from \textit{in-the-wild} Kinetics to robotic-centric RGB-Stacking---to the high-resolution pixel alignment enabled by our warping-based architecture.

Beyond point-tracking accuracy, \method also delivers superior occlusion classification.
On average, we improve OA by {3.0} over AllTracker (Kub) and by {1.8} over AllTracker (Kub+Mix), with the largest gap on DAVIS (+4.3).
We hypothesize that this advantage arises because our head operates directly on image features rather than compressing appearance into correlation scores when building cost volumes: dot-product similarity can discard channel-wise cues that are predictive of visibility at boundaries, whereas warping-indexed features preserve such information for the occlusion head.

\subsubsection{Qualitative Results}

\Cref{fig:bmx-occlusion} visualizes a challenging BMX sequence with a long, full occlusion in the middle frames.
We compare DELTA, AllTracker, and our method under the same evaluation protocol.
Before the occlusion, all three methods roughly follow the rider; however, as the cyclist passes behind the foreground object, DELTA rapidly loses the target and fails to re-acquire it afterward, leading to large swaths of missing or mislocalized tracks.
AllTracker retains partial correspondence but exhibits visible drift and fragmentation: trajectories spread into the background and then snap back with an offset, producing inconsistent motion on the rider's limbs and the bicycle outline.
In contrast, our tracker maintains a coherent hypothesis throughout the occlusion and cleanly locks back onto the rider once visibility returns.
The re-acquired tracks align tightly with object boundaries, including thin structures such as the handlebar and wheel, and remain stable through the landing.

\input{tables/flow_main}    %
\input{figs/flow_results}   %
\input{figs/acc_mag_plot}   %

\subsection{Optical Flow}%
\label{sec:exp-flow}

We now evaluate how well \method transfers to optical flow estimation by simply treating a frame pair as a two-frame ``video''.
We consider three standard benchmarks:
MPI-Sintel~\cite{butler12a-naturalistic},
KITTI-2015~\cite{geiger12are-we-ready}, and
Spring~\cite{mehl23spring:}, which cover cinematic synthetic sequences with complex non-rigid motion (Sintel), real-world driving scenes with strong camera and background motion (KITTI), and high-quality rendered videos with fine structures and realistic motion (Spring).
All predictions are produced by the same model used for point tracking, without any additional training on optical-flow data.

For each benchmark, we follow the standard evaluation protocols.
All inputs are resized to a resolution of $336\times 560$.
We treat each frame pair as a two-frame video and run \method with exactly the same configuration as in our video setting, without any flow-specific tuning or changes to hyperparameters.
We report End-Point Error (EPE) on all datasets, Fl-all on KITTI-2015, and the 1px error rate on Spring.

\input{tables/ablations}   %

\subsubsection{Quantitative Results}

\Cref{tab:flow_results} reports zero-shot optical flow accuracy on Sintel, KITTI-2015 (train), and Spring (val), where predictions are produced by the \emph{same} model used for point tracking without any flow-specific training.
On Sintel, our warping-only tracker attains {0.78} EPE on the Clean split and {1.48} EPE on the Final split, improving over the next best specialized flow model (0.94/1.86) by {17\%}/{20\%} relative, respectively.
These gains suggest that \method recovers pixel-accurate motion and preserves thin structures, yielding state-of-the-art zero-shot performance on both Sintel splits.

On KITTI-2015, our model achieves {1.04} EPE and {4.87}\% Fl-all, outperforming RAFT, SEA-RAFT, and both WAFT variants.
Relative to the strongest prior (WAFT twins-a2), this is a {9.6\%} reduction in EPE and a {7.9\%} reduction in Fl-all.
This suggests an advantage under wide-baseline motions and in texture-poor regions common in driving scenes, where cost-volume methods either lower the feature resolution or restrict the search radius; our warping-based architecture allows the head to index at higher spatial resolution and convert additional input detail directly into accuracy.

On Spring, our tracker remains competitive despite not being trained on flow data, obtaining {0.17} EPE and {0.75}\% 1px.
Specialized WAFT models achieve slightly smaller absolute errors (0.11--0.13 EPE and 0.70--0.74\% 1px), but our zero-shot results are within {0.06} EPE and {0.05}\% 1px of the best entries.
We note that Spring operates in a very low-motion regime: the average flow magnitude is only $3.5$ pixels, compared to $13.4$ pixels on Sintel and $27.8$ pixels on KITTI-2015.
In this regime, all methods achieve extremely small errors, and the gaps are close to the scale of noise, making it difficult to draw strong conclusions from Spring alone.

\paragraph{Accuracy vs.~motion scales.}

To understand how our warping-based head behaves across motion scales, we stratify errors by flow magnitude and report EPE (averaged across all pixels in a flow-magnitude bin) as a function of motion in \cref{fig:acc_mag}.
Compared to SEA-RAFT, our method not only attains consistently lower EPE across all bins, but also exhibits a substantially flatter linear trend (slope $0.16$ vs.~$0.30$).
The relative improvement becomes most pronounced for large displacements: in the highest-motion bin, our EPE is roughly {46\%} lower than SEA-RAFT, compared to an average improvement of about {22.5\%}.
This analysis suggests that, despite lacking an explicit cost volume, our warping-only design does \emph{not} struggle with large motions; rather, high-resolution local warping appears more robust when two matching features are far apart, whereas correlation volumes over large search regions may produce noisy or ambiguous responses.

\subsubsection{Qualitative Results}

\Cref{fig:sintel_flow_predictions} compares our optical flow predictions with the ground-truth flows.
Our method yields smooth, accurate motion with sharp boundaries even in difficult scenarios, including large displacements, occlusions, and cluttered backgrounds.

\input{figs/iteration_plots}           %

\subsection{Ablation Studies}%
\label{sec:ablation}

In \cref{tab:ablations} and \cref{fig:iteration_plots}, we ablate our main design choices:

\paragraph{Backbone choice.}
\Cref{tab:ablation:num_train_tracks} compares interchangeable backbones under the same warping-only head.
The ConvNet-based backbone used by CoTracker performs clearly worse, while ViT (we use the patch embedding network of VGGT) and Pi$^3$ works reasonably well.
VGGT achieves the best \davgvis on all three datasets, improving over Pi$^3$ by +1.7 \davgvis on average, showing that our head benefits directly from stronger video backbones.

\paragraph{Upsampling for high-resolution indexing.}
\Cref{tab:ablation:num_test_tracks} compares ways to lift low-resolution features for high-resolution warping.
Removing the upsampler hurts performance; bilinear upsampling helps; LoftUp-style cross-attention helps slightly more; and the DPT upsampler performs best, with gains of +5.5 \davgvis on DAVIS and +3.4 \davgvis on RoboTAP over no upsampler.

\paragraph{Indexing resolution vs.~patch size.}
In \Cref{tab:ablation:track_type}, we vary the indexing resolution (feature stride vs.~input) while adjusting the transformer patch size to roughly match compute.
Even with compensating patch sizes, higher indexing resolution (1/2 stride) clearly outperforms coarser settings (1/4--1/16), reaching 78.0 \davgvis on DAVIS and 83.4 on RoboTAP\@.
This supports prioritizing high-resolution indexing.

\paragraph{Spatial vs.~spatio-temporal tracker network.}
\Cref{tab:ablation:temporal_modeling} compares a purely spatial image transformer (ViT (img.)) with our spatio-temporal transformer that adds temporal attention.
Temporal attention yields large gains on long sequences (RGB-Stacking and RoboTAP, up to 600 frames), improving \davgvis by +11.7 and +11.2, respectively.

\paragraph{Iterative refinement.}
\Cref{tab:ablation:attention_layers} contrasts a single-pass head with our iterative refinement scheme.
Multiple refinement steps in the warping head significantly improve performance, by +6.6 \davgvis on DAVIS and +4.0 \davgvis on RoboTAP.

\paragraph{Head design.}
Finally, \Cref{tab:ablation:layer_pos} evaluates tracker head designs.
A non-warping variant that always queries the original features performs dramatically worse ($-23.4$ \davgvis on DAVIS, $-9.6$ \davgvis on RoboTAP), highlighting the importance of explicit warping in \method.
\method also compares favourably to an AllTracker head on the same backbone.
We hypothesize that this is because our warping-only head uses backbone features directly and can operate at higher resolution, whereas correlation volumes are expensive and thus confined to coarser grids.

\paragraph{Refinement steps.}
\Cref{fig:iteration_plots} shows that increasing the refinement steps $K$ consistently boosts performance, with the largest gain from $K=1$ to $K=2$ and diminishing returns thereafter.
Metrics stabilize around $K=5$, which we adopt as our default.

%% file: figs/bmx_comparison.tex
\begin{figure*}[t]
\centering
\includegraphics[width=\linewidth]{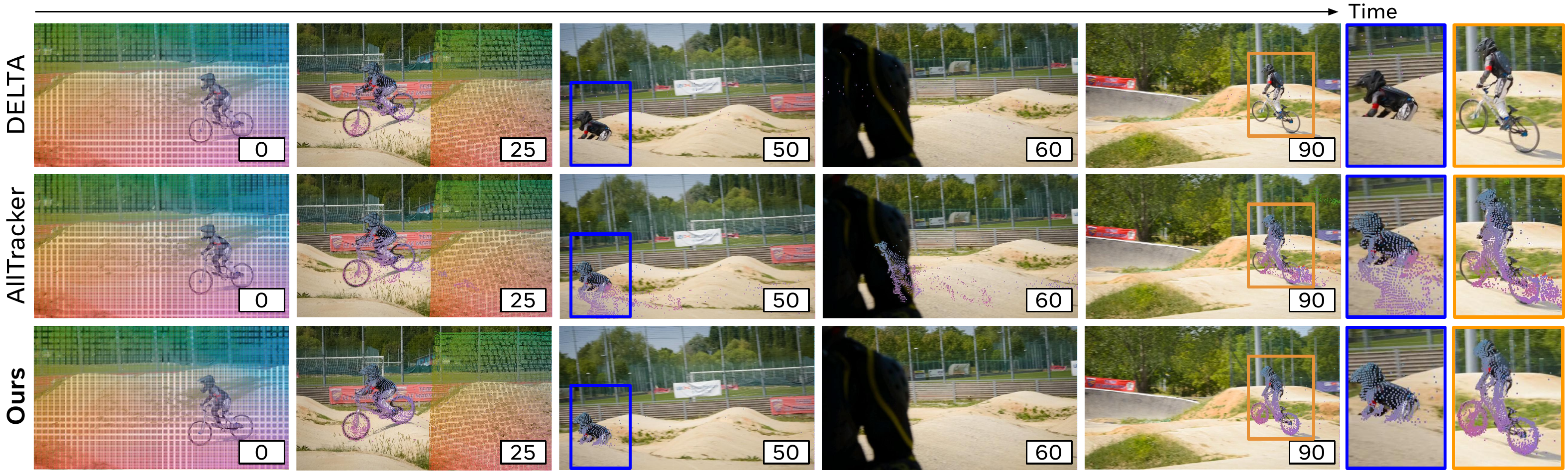}
\vspace{-0.3in}
\caption{\textbf{Tracking through a challenging BMX sequence with a full occlusion in the middle frames. }
Rows compare {DELTA}, {AllTracker}, and {Ours}. 
Our method maintains a consistent track before, during, and after occlusion, whereas {DELTA} loses the target and fails to recover and {AllTracker} exhibits noticeable drift and fragmentation. 
Our warp-based indexing queries features at high resolution, preserving fine details and enabling accurate localization after occlusion. Numbers in lower-right boxes indicate frame numbers.}%
\vspace{-1em}
\label{fig:bmx-occlusion}
\end{figure*}

%% file: tables/flow_main.tex
\begin{table}[t]
  \centering
  \setlength{\tabcolsep}{2.4pt}
  \footnotesize
  \begin{tabular}{lcccccc}
    \toprule
    & \multicolumn{2}{c}{\textbf{Sintel}} & \multicolumn{2}{c}{\textbf{KITTI (train)}} & \multicolumn{2}{c}{\textbf{Spring (val)}} \\
    \cmidrule(lr){2-3} \cmidrule(lr){4-5} \cmidrule(lr){6-7}
    \textbf{Method} & Clean$\downarrow$ & Final$\downarrow$ & EPE$\downarrow$ & Fl-all$\downarrow$ & EPE$\downarrow$ & 1px$\downarrow$ \\
    \midrule
    \multicolumn{7}{l}{\textit{Specialized Optical Flow model}} \\
    \midrule
    RAFT~\cite{teed20raft:} & 1.15 & 1.86 & 1.53 & 7.81 & 0.22 & 1.79 \\
    SEA-RAFT~\cite{wang2024sea} (M) & 0.97 & 1.96 & 1.60 & 8.26 & 0.21 & 1.44 \\
    WAFT~\cite{wang25waft:} (twins-a2) & 0.94 & 2.09 & 1.15 & 5.29 & \textbf{0.11} & 0.74 \\
    WAFT~\cite{wang25waft:} (Dv3-a2) & 1.01 & 1.86 & 1.35 & 6.41 & 0.13 & \textbf{0.70} \\
    \midrule
    \multicolumn{7}{l}{\textit{Dense Point Tracker}} \\
    \midrule
    \textbf{\method} & \textbf{0.78} & \textbf{1.48} & \textbf{1.04} & \textbf{4.87} & 0.17 & 0.75 \\
    \bottomrule
  \end{tabular}
  \vspace{-1em}
  \caption{\textbf{Zero-shot optical flow results} on three benchmarks. Our predictions are produced by the \emph{same model used for point tracking}, and the model was \emph{not trained} on any optical-flow datasets, including Sintel. Despite this, \method  shows strong zero-shot transfer from tracking to optical flow.}
  \vspace{-1em}
  \label{tab:flow_results}
\end{table}

%% file: figs/flow_results.tex
\begin{figure*}[t]
\centering
\includegraphics[width=\textwidth]{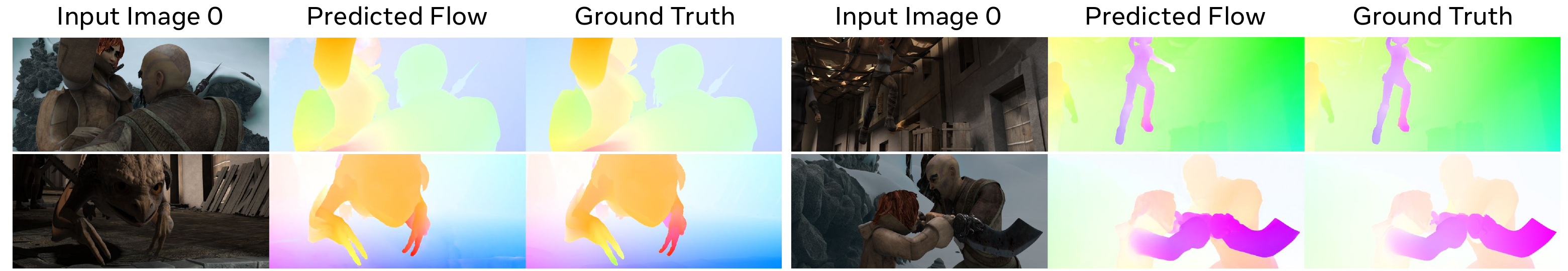}
\vspace{-2em}
\caption{\textbf{Optical flow predictions on MPI Sintel} using the \emph{same} model as our point-tracking results. 
The predicted flows closely match the ground truth even in difficult scenarios—\emph{large motion}, \emph{occlusions}, and \emph{background clutter}. 
Note that the model was \emph{not trained} on any optical-flow datasets, including Sintel.}%
\vspace{-2em}
\label{fig:sintel_flow_predictions}
\end{figure*}

%% file: figs/acc_mag_plot.tex
\begin{figure}[h!]
    \centering
    \includegraphics[width=\linewidth]{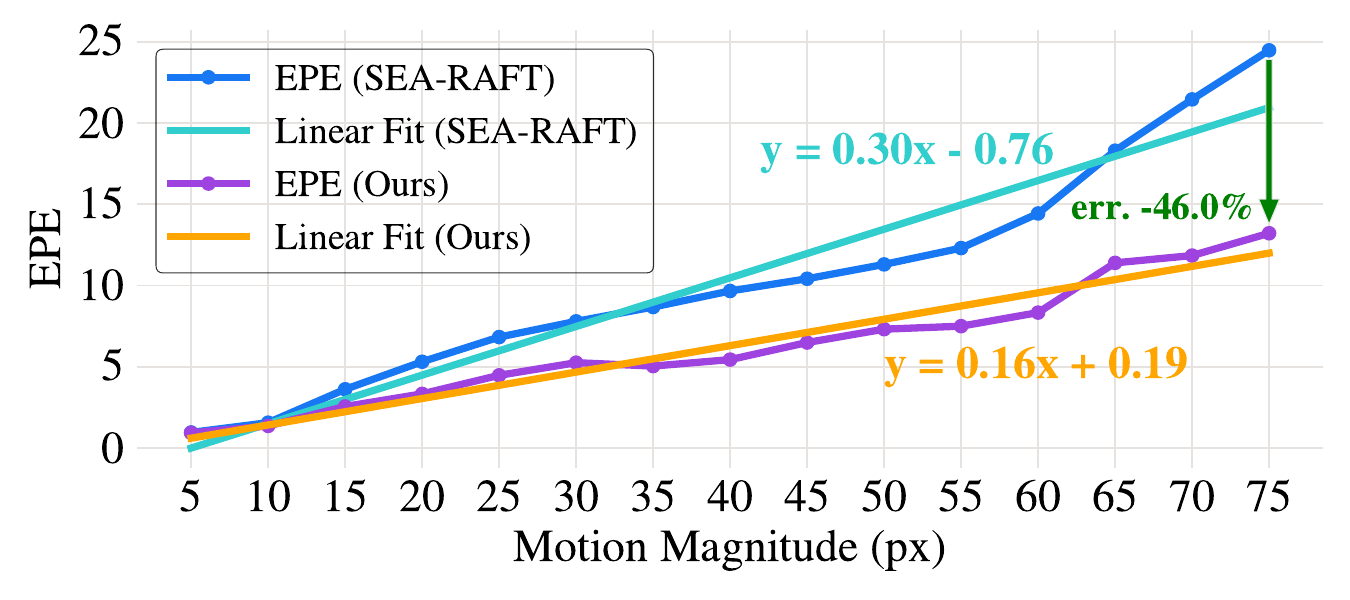}
    \vspace{-2em}
    \caption{\textbf{EPE vs. motion magnitude} for SEA-RAFT and \method, evaluated  on all pixels in Sintel. Our method yields lower error across all motion bins and a flatter increase with motion (slope 0.16 vs. 0.30), with the largest gains at high displacements ($46\%$ lower EPE), indicating stronger robustness to large motion.}
    \vspace{-1em}
    \label{fig:acc_mag}
\end{figure}

%% file: tables/ablations.tex
\begin{table*}[!ht]
\centering

\subfloat[\textbf{Backbone}: VGGT backbone yields best results across datasets, whereas Pi$^3$ also produce reasonable results.\label{tab:ablation:num_train_tracks}]{
\tablestyle{5pt}{1}
\begin{tabular}{x{38}x{25}x{25}x{25}}\toprule
\multicolumn{1}{c}{Model}  & DAVIS$\uparrow$ & RGB$\uparrow$ & Rob.$\uparrow$ \\
\midrule
CoTracker~\cite{karaev25cotracker3}  & 62.9 & 77.1 & 68.3  \\
ViT~\cite{wang25vggt}  & 75.7 & 89.9  &  80.9 \\
Pi$^3$~\cite{wang2025pi}  & 75.3  & 91.1 & 82.6 \\
\rowcolor[gray]{0.9}
VGGT~\cite{wang25vggt} & \textbf{78.0} & \textbf{92.8} & \textbf{83.4}\\
\bottomrule
\end{tabular}}%
\hspace{3mm}
\subfloat[\textbf{Upsampler}: DPT upsampler gives strongest \davgvis, outperforming bilinear and a Loftup-style upsampler for high-resolution indexing.\label{tab:ablation:num_test_tracks}]{
\tablestyle{5pt}{1}
\begin{tabular}{x{38}x{25}x{25}x{25}}\toprule
\multicolumn{1}{c}{Type}  & DAVIS$\uparrow$ & RGB$\uparrow$ & Rob.$\uparrow$ \\
\midrule
None  & 72.5 & 90.3 & 80.0  \\
Bilinear  & 76.8 & 91.9 & 82.0 \\
Loftup~\cite{huang2025loftup} & 77.4 & 90.8 & 81.9  \\
\rowcolor[gray]{0.9}
DPT~\cite{ranftl21vision} & \textbf{78.0} & \textbf{92.8} & \textbf{83.4} \\
\bottomrule
\end{tabular}}%
\hspace{3mm}
\vspace{1mm}
\subfloat[\textbf{Resolution–patch-size tradeoff.} A finer indexing stride (1/2) achieves the best results, outperforming coarser strides.
\label{tab:ablation:track_type}]{
\tablestyle{5pt}{1}
\begin{tabular}{x{38}x{25}x{25}x{25}}\toprule
\multicolumn{1}{c}{Ratio, Patch}  & DAVIS$\uparrow$ & RGB$\uparrow$ & Rob.$\uparrow$ \\
\midrule
1/16, 1x1 & 70.9 & 89.8 & 80.1  \\
1/8, 1x1  & 74.6 & 91.6 & 82.3  \\
1/4, 2x2  & 75.0 &  91.6 & 81.8  \\
\rowcolor[gray]{0.9}
1/2, 4x4 & \textbf{78.0} & \textbf{92.8} & \textbf{83.4} \\
\bottomrule
\end{tabular}}%
\hspace{3mm}
\subfloat[\textbf{Tracker network structure}: Adding  temporal attention in the head greatly boosts performance on long RGB and RoboTAP videos.\label{tab:ablation:temporal_modeling}]{
\tablestyle{5pt}{1}
\begin{tabular}{x{38}x{25}x{25}x{25}}\toprule
\multicolumn{1}{c}{Model}  & DAVIS$\uparrow$ & RGB$\uparrow$ & Rob.$\uparrow$ \\
\midrule
ViT (img.) & 74.7 & 81.1 & 72.6 \\
\rowcolor[gray]{0.9}
Ours (vid.) & \textbf{78.0} & \textbf{92.8} & \textbf{83.4} \\
$\Delta$ & +3.3 & +11.7 & +11.2 \\
\bottomrule
\end{tabular}}%
\hspace{3mm}
\subfloat[\textbf{Iterative refinement}: Multi-step warping updates significantly outperform single-pass refinement on all three benchmarks.\label{tab:ablation:attention_layers}]{
\tablestyle{5pt}{1}
\begin{tabular}{x{38}x{25}x{25}x{25}}\toprule
\multicolumn{1}{c}{Iterative?}  & DAVIS$\uparrow$ & RGB$\uparrow$ & Rob.$\uparrow$ \\
\midrule
No  & 71.4 & 90.0 & 79.4 \\
\rowcolor[gray]{0.9}
Yes & \textbf{78.0} & \textbf{92.8} & \textbf{83.4} \\
$\Delta$ & +6.6 & +2.8 & +4.0 \\
\bottomrule
\end{tabular}}%
\hspace{3mm}
\subfloat[\textbf{Head design}: Our full warping-only head significantly outperforms a non-warping variant and also the AllTracker  head.\label{tab:ablation:layer_pos}]{
\tablestyle{5pt}{1}
\begin{tabular}{x{48}x{23}x{23}x{19}}\toprule
\multicolumn{1}{c}{Head}  & DAVIS$\uparrow$ & RGB$\uparrow$ & Rob.$\uparrow$ \\
\midrule
AllTracker  & 72.0 &  89.5 &  80.6 \\
Ours(no warp)  & 54.6 & 85.5  &  73.8 \\
\rowcolor[gray]{0.9}
Ours& \textbf{78.0} & \textbf{92.8} & \textbf{83.4} \\
\bottomrule
\end{tabular}}%
\hspace{3mm}
\vspace{-0.1in}
\caption{\textbf{Ablation study for \method.} Each table isolates the effect of a specific design choice. Results are reported using \davgvis.}
\vspace{-0.1in}

\label{tab:ablations}
\end{table*}

%% file: figs/iteration_plots.tex
\begin{figure}[t]
\centering
\includegraphics[trim=0 35 0 0,clip,width=0.95\linewidth]{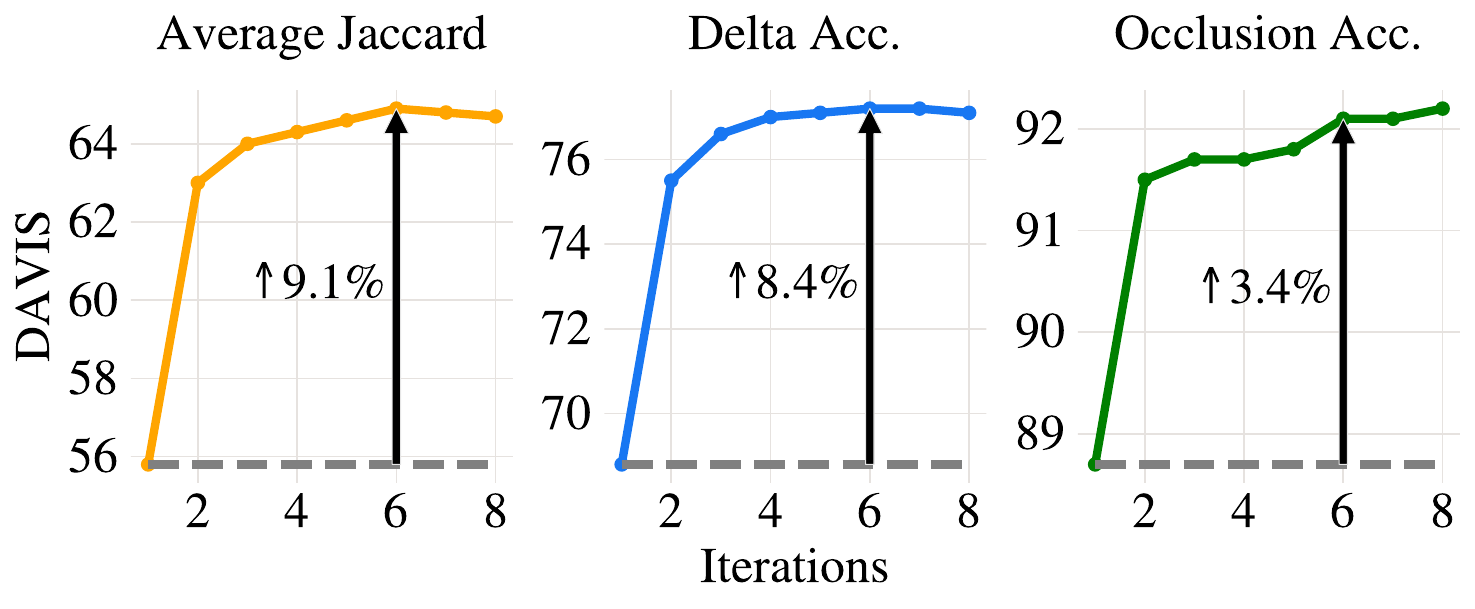}
\includegraphics[trim=0 0 0 29,clip,width=0.95\linewidth]{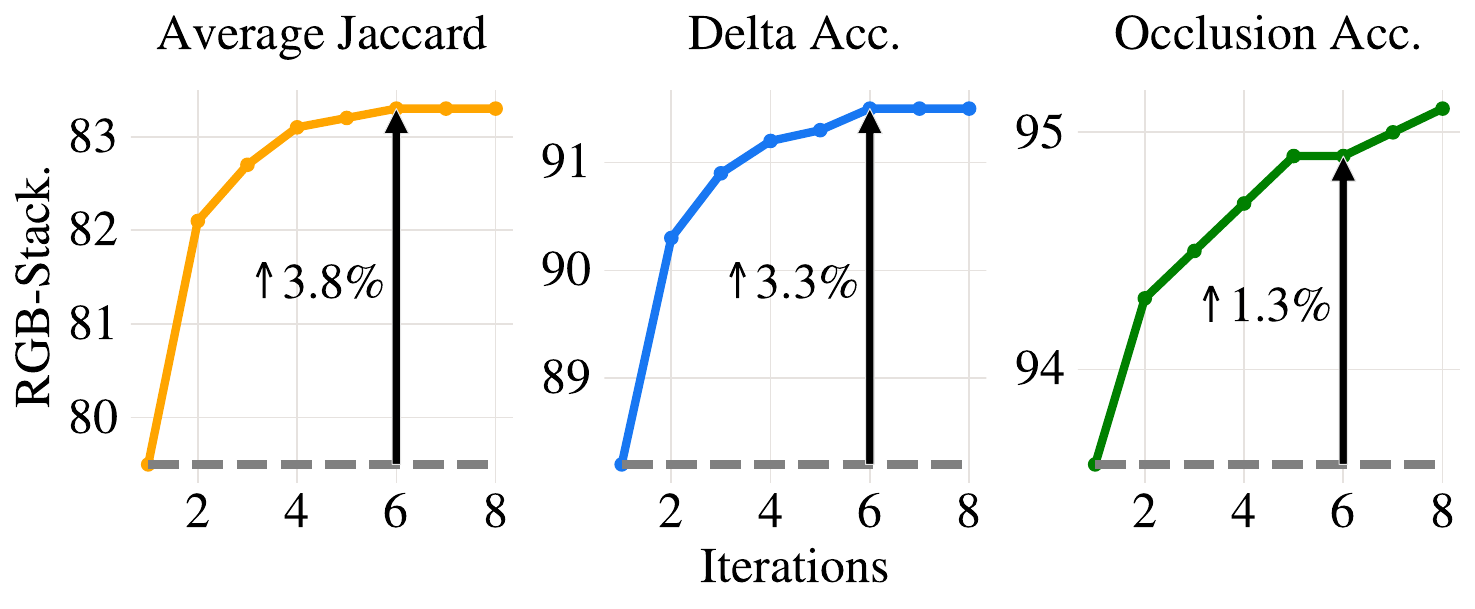}
\vspace{-1em}
\caption{\textbf{Ablation on test-time iterations.}
We vary the number of refinement steps $K$.
Performance improves significantly from $K=1$ to $K=2$, then gradually saturates around $K=5$--$6$, with OA yielding modest additional gains.}%
\vspace{-0.25in}
\label{fig:iteration_plots}
\end{figure}

%% file: sec/5_conclusions.tex
\section{Limitations}
CoWTracker has several limitations that suggest future research directions. Our method may fail under extreme viewpoint changes, long-range full occlusions, or severe specularities. While the iterative warping head is lightweight, overall throughput is heavily dependent on the chosen backbone, such as VGGT. Nonetheless, the marginal overhead of CoWTracker remains small, and the benefits of higher-resolution feature indexing persist even when using more efficient backbones. Another constraint is the quadratic complexity of the VGGT backbone with respect to video length, which requires processing longer clips in chunks. Furthermore, the current iterative refinement process tends to saturate after five to six steps, indicating a ceiling on the model's self-correction capability. Finally, the model is currently trained exclusively on synthetic Kubric data. Diverse real-world data remains significantly underleveraged; incorporating natural videos could improve robustness to lighting and noise patterns that are difficult to simulate.
\section{Conclusions}%
\label{sec:conclusions}

We presented \method, a dense point tracker that replaces cost volumes with iterative warping-based refinement. This simple, warp-based head scales linearly with the spatial resolution and attains state-of-the-art tracking on TAP-Vid and RoboTAP benchmarks while transferring competitively to optical flow estimation without flow-specific training.
We hope these results will encourage revisiting dense correspondence architecture with simple, warp-centric designs that bridge tracking and optical flow.

%% file: sec/X_suppl.tex
\clearpage
\setcounter{page}{1}
\maketitlesupplementary

\section{Implementation Details}
\label{sec:sup_imp}
\subsection{Pseudo-code of \method}

In \Cref{fig:pseudocode}, we provide a minimal PyTorch-style pseudocode of the proposed WarpTracker update loop. First, we extract features for all video frames with a standard backbone and replicate the anchor-frame features for all queries. We then initialize the track field and the hidden state once. After this setup, the algorithm performs a very simple iterative update: at each of the \(K\) iterations, we (i) warp the features according to the current tracks, (ii) concatenate the original features, warped features, tracks, and hidden state, (iii) update the hidden state, and (iv) refine the tracks with a small linear head.
Importantly, no correlation or cost volume is constructed at any point in this loop; instead, we operate directly on the raw backbone features of each frame. This design makes the method both easy to implement and straightforward to integrate into existing codebases. Please refer to the main paper for a detailed description of each component.

\section{Extended Quantitative Results}
\label{sec:sup_quant}

In this section, we provide expanded quantitative comparisons that complement the results presented in the main paper.

\subsection{Extended Evaluation on the AllTracker Benchmark Suite}

In addition to the datasets considered in the main paper, we further evaluate our method on the extended benchmark suite introduced by AllTracker~\cite{harley25alltracker:}. This suite augments standard point-tracking benchmarks with three additional datasets (DriveTrack, EgoPoints, and Horse10) covering challenging driving scenes, egocentric videos, and articulated animal motion, respectively. We follow exactly the same evaluation protocol and dataset splits as AllTracker, and report the $\delta_{\text{avg}}$ metric averaged over all sequences.

Table~\ref{tab:supp_alltracker_extended} compares \method with recent sparse trackers, dense trackers, and optical-flow models. Our method achieves the best overall performance on this suite, obtaining an average $\delta_{\text{avg}}$ of {73.6}, which improves upon the strongest AllTracker  by {+2.3} points. \method is best or tied-best on all individual datasets, including the three newly added benchmarks, demonstrating that \method generalizes well across diverse motion patterns and scene types.

\input{figs/fig_pseudocode}
\input{tables/tracking_others}

\subsection{Runtime Analysis}

We benchmark the runtime of \method on a single NVIDIA H100 GPU with FlashAttention-3~\cite{shah24flashattention-3:}. For all results we use $K{=}5$ refinement steps, and input resolution $336 \times 560$.

Table~\ref{tab:runtime} reports end-to-end runtime as a function of video length, together with a breakdown into backbone and tracker cost, as well as the resulting point and frame throughput. The tracker itself scales approximately linearly with the number of frames and contributes a smaller fraction of the total runtime for longer sequences. For shorter clips ($\leq 40$ frames) \method tracks around $6$--$7$ million points per second and runs at over $30$ frames per second. For longer videos throughput gradually decreases but remains around $3.5$ million points per second and nearly $20$ frames per second.

The main runtime growth for long sequences stems from the VGGT backbone, whose complexity scales quadratically with video length. We note that (i) the backbone is essentially orthogonal to our tracker, meaning that any future advances in backbone architecture---including more efficient designs---will directly improve end-to-end speed, and (ii) because the backbone is a general-purpose 3D representation model, WarpTracker effectively acts as a lightweight add-on that can be plugged into existing architectures used for other tasks, delivering the new functionality with very little extra runtime. In practice, one can also process long videos in shorter chunks to avoid the quadratic cost of VGGT, trading only about $1\%$ in $\delta$ accuracy for a substantial runtime reduction. 

\subsection{Memory Analysis}
\input{tables/runtime}

Figure~\ref{fig:supp_memory} reports the memory required to store
either a cost volume or a pair of feature maps (the alternative used by our
cost-volume-free warping based head). Cost volumes grow roughly $16\times$ in memory whenever the image
resolution doubles, because both the number of spatial locations and the
search window increase. As a result, using stride~$1/2$ features becomes
prohibitively expensive (hundreds of gigabytes in our setting), and even
stride~$1/4$ is costly. This is why most prior dense trackers operate at
stride~$1/8$, trading away high-frequency details and often degrading tracking
accuracy, especially around thin structures and object boundaries.

Our method avoids cost volumes entirely and operates directly on feature maps
using a warping-based head. This design removes the quadratic dependence on
the spatial search window and yields much more favorable memory scaling with
resolution. As shown in \cref{fig:supp_memory}, our head can run
comfortably on stride~$1/2$ features, which only takes a few gigabytes of memory to store,
enabling high-resolution tracking without sacrificing practicality.

\input{supp_figs/memory_cost}

\section{Extended Qualitative Results}
\label{sec:extended_quali_results}
\input{supp_figs/trackcomparison1}
\input{supp_figs/trackothers}
\input{supp_figs/flow_results_supp}

In this section, we provide additional qualitative visualizations to further demonstrate the robustness and generalization capabilities of our proposed method.

\subsection{Comparison to Existing Methods}
We first evaluate our model on highly challenging sequences characterized by non-rigid motion and severe occlusions and compare with existing methods. As illustrated in Figure~\ref{fig:supp_track_compare}, the top row (``Dive-in'') presents a scenario where a person enters the water, causing significant non-rigid deformation and drastic appearance changes. While baseline methods like DELTA and AllTracker lose the target or drift significantly, our method maintains a consistent track. 

Furthermore, the bottom row of Figure~\ref{fig:supp_track_compare} highlights a scenario with rapid camera motion where the target object frequently exits the camera frame. This sequence also necessitates tracking a small object (a drone). Our method successfully handles these re-entry scenarios and accurately tracks the small target, a capability we attribute to our utilization of high-resolution feature maps which preserve fine-grained spatial details.

\subsection{Generalization Across Domains}
To assess the universality of our tracking model, we test \method on videos from three distinct domains without fine-tuning. Figure~\ref{fig:supp_track_generalize} displays qualitative results on:
\begin{itemize}
    \item \textbf{Ego-centric videos} (top row), where camera motion is erratic and hand-object interaction is frequent.
    \item \textbf{Self-driving scenes} (middle row), involving dynamic street environments.
    \item \textbf{Robotics manipulation} (bottom row), requiring precise tracking of manipulated objects.
\end{itemize}
Notably, our method exhibits strong temporal stability, successfully maintaining accurate tracks over very long sequences (up to 600 frames), as demonstrated in the ego-centric and robotics examples.

\subsection{Zero-Shot Optical Flow Estimation}
Although our architecture is designed for tracking videos, it naturally produces optical flow by treating an image pair as a 2-frame ``video''. In the main paper, we presented results on the Sintel dataset. Here, in Figure~\ref{fig:supp_flow}, we extend this analysis to the Spring~\cite{mehl23spring:} (top two rows) and KITTI~\cite{geiger12are-we-ready} (bottom two rows) datasets. 

Qualitatively, our predicted flow exhibits sharp motion boundaries and accurate alignment with the ground truth. Crucially, we emphasize that our model is \textbf{not} trained on any optical flow datasets (including Sintel, Spring, or KITTI). These results are achieved using the exact same tracking weights, highlighting the generality of motion understanding ability embedded in our model.

%% file: figs/fig_pseudocode.tex
\begin{figure}[t]
  \centering
  \includegraphics[width=0.99\linewidth]{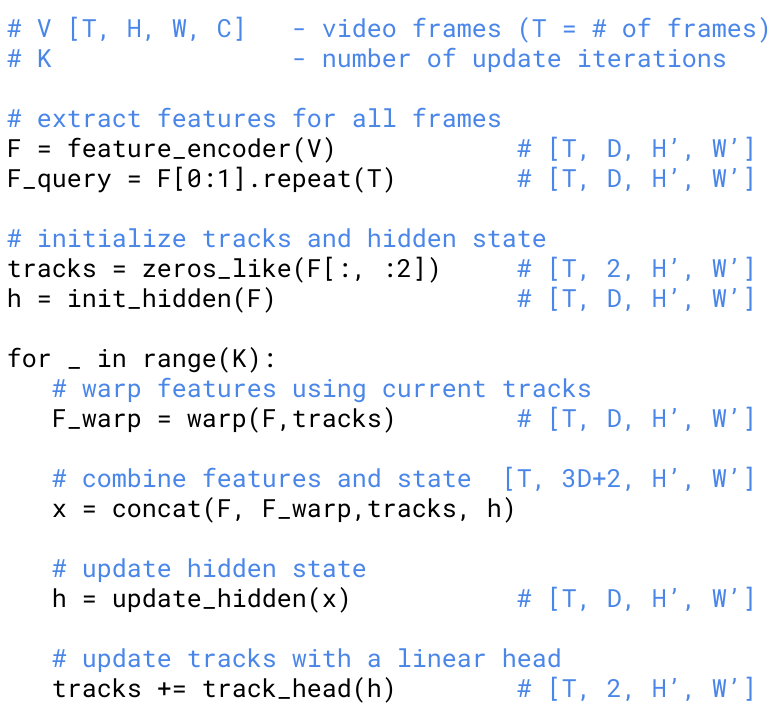}
  \caption{\textbf{Pseudocode for \method algorithm:} our model iteratively refines track prediction across all frames using repeated warping and updates, and as shown, is also simple to implement.
}%
\label{fig:pseudocode}
\end{figure}

%% file: tables/tracking_others.tex
\begin{table}[t]
\setlength{\tabcolsep}{1.9pt}
\footnotesize
\centering
\vspace{-0.5em}
\begin{tabular}{llccccccc>{\columncolor[gray]{0.9}}c}
\toprule
Method & Data & Dav. & Dri. & Ego. & Hor. & Kin. & Rgb. & Rob. & Avg. \\
\midrule
\multicolumn{9}{l}{\textit{Optical Flow Models*}} \\
\midrule
AccFlow~\citep{wu2023accflow} & Flow & 23.5 & 26.4 & 4.0 & 12.1 & 38.8 & 63.2 & 57.9 & 32.3 \\
RAFT~\citep{teed20raft:} & Flow & 48.5 & 44.8 & 41.0 & 27.8 & 64.3 & 82.8 & 72.2 & 54.5 \\
SEA-RAFT~\citep{wang2024sea} & Flow & 48.7 & 49.4 & 44.0 & 33.1 & 64.3 & 85.7 & 67.6 & 56.1 \\
\midrule
\multicolumn{9}{l}{\textit{Sparse Trackers}} \\
\midrule
PIPs++~\cite{harley22particle} & PO & 62.5 & 51.3 & 38.5 & 21.4 & 64.2 & 70.4 & 73.4 & 54.5 \\
CoTracker2~\cite{karaev24cotracker} & Kub & 70.9 & 67.8 & 43.2 & 33.9 & 65.8 & 73.4 & 73.0 & 61.1 \\
LocoTrack~\cite{cho24local} & Kub & 68.0 & 66.5 & 58.4 & 48.9 & 70.0 & 80.3 & 76.9 & 67.0 \\
BootsTAPIR~\cite{doersch24bootstap:} & Kub+ & 67.9 & 66.9 & 56.8 & 48.8 & 70.6 & 81.0 & 78.2 & 67.2 \\
CoTracker3-Kub~\cite{karaev24cotracker3:} & Kub & 77.4 & {69.8} & 58.0 & 47.5 & 70.6 & 83.4 & 77.2 & 69.1 \\
CoTracker3~\cite{karaev24cotracker3:} & Kub+ & 77.1 & {69.8} & 60.4 & 47.1 & 71.8 & 84.2 & 81.6 & 70.3 \\
\midrule
\multicolumn{9}{l}{\textit{Dense Trackers}} \\
\midrule
DELTA~\cite{ngo24delta:} & Kub & 75.3 & 67.8 & 40.3 & 41.8 & 66.5 & 83.0 & 74.8 & 64.2 \\
AllTracker-Kub~\cite{harley25alltracker:} & Kub & 75.2 & 66.1 & 60.3 & 49.0 & 71.3 & 90.1 & 82.2 & 70.6 \\
AllTracker~\cite{harley25alltracker:} & Kub+ & 76.3 & 65.8 & 62.5 & 49.0 & 72.3 & 90.0 & \textbf{83.4} & 71.3 \\
\method & Kub & \textbf{78.0} & \textbf{70.7} & \textbf{64.4} & \textbf{52.7} & \textbf{73.1} & \textbf{92.8} & \textbf{83.4} & \textbf{73.6} \\
\bottomrule
\end{tabular}
\caption{\textbf{Comparison against state-of-the-art point trackers and optical-flow baselines (incl. extra eval datasets used in AllTracker~\cite{harley25alltracker:}).} Baseline numbers are taken from~\cite{harley25alltracker:}. *Optical-flow models are assessed \emph{zero-shot} on the tracking task without any task-specific fine-tuning, and understandably yield lower results.} 
\label{tab:supp_alltracker_extended}
\end{table}

%% file: tables/runtime.tex
\begin{table}[t]
  \centering
  \small
  \setlength{\tabcolsep}{2pt}  
  \begin{tabular}{@{}lcccccccc@{}}
    \toprule
    \bfseries Input Frames 
    & \bfseries $2^*$ 
    & \bfseries $10$ 
    & \bfseries $20$ 
    & \bfseries $40$ 
    & \bfseries $60$ 
    & \bfseries $80$ 
    & \bfseries $100$
    & \bfseries $200$ \\
    \midrule
    \textbf{All} 
      & 0.07
      & 0.27
      & 0.57
      & 1.37
      & 2.42
      & 3.75
      & 5.35
      & 10.9
      \\
    \textbf{- Backbone} \emph{(VGGT)}
      & 0.02
      & 0.10
      & 0.25
      & 0.76
      & 1.51
      & 2.54
      & 3.85
      & 7.89
      \\
    \textbf{- Tracker} 
      & 0.05
      & 0.17
      & 0.32
      & 0.61
      & 0.91
      & 1.21
      & 1.50
      & 3.01
      \\
    \textbf{Point/sec ($\times 10^6$)} 
      & 5.37
      & 6.96
      & 6.60
      & 5.49
      & 4.66
      & 4.01
      & 3.51
      & 3.45
      \\
    \textbf{Frame/sec} 
      & 28.6
      & 37.0
      & 35.1
      & 29.2
      & 24.8
      & 21.3
      & 18.7
      & 18.4
      \\
        \bottomrule
  \end{tabular}
  \caption{\textbf{\method runtime vs.\ video length. }
  We report \emph{Runtime (in seconds, all and by components, \da~better)} and \emph{Point / Frame per Second (\ua~better)}. *equivalent to optical flow runtime.}
  \label{tab:runtime}
\end{table}

%% file: supp_figs/memory_cost.tex
\begin{figure}[t]
\centering
\includegraphics[width=0.99\linewidth]{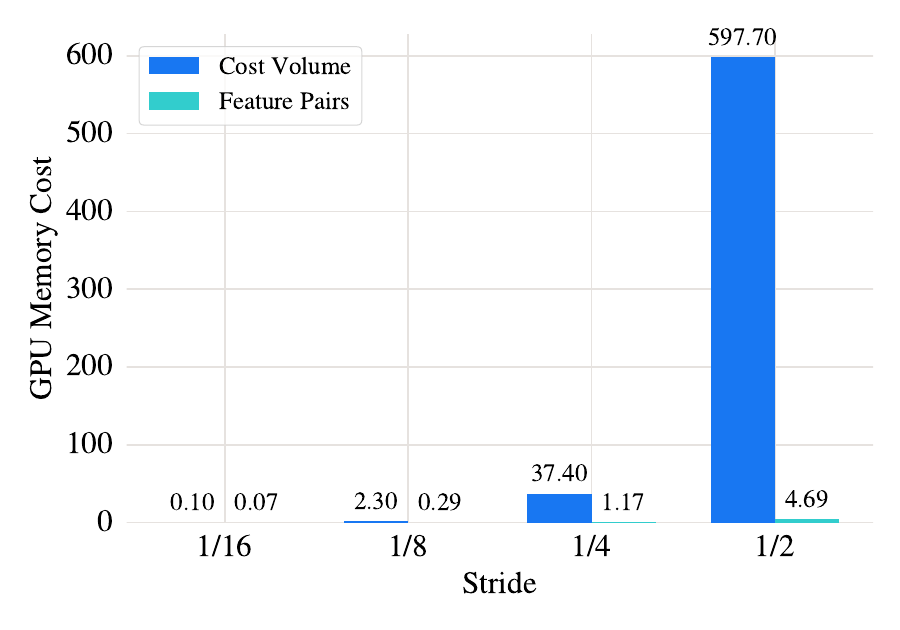}
\caption{\textbf{Cost volumes are prohibitively expensive at high resolution.}
Memory grows by $\sim16\times$ whenever the image size doubles, forcing most methods to use stride~1/8 and sacrifice accuracy. Compared to cost volumes, feature pairs are far more memory efficient. Numbers computed using 100-frame video length, 384x512 resolution, and 16-bit precision.}
\label{fig:supp_memory}
\end{figure}

%% file: supp_figs/trackcomparison1.tex
\begin{figure*}[t]
\centering
\includegraphics[width=0.99\linewidth]{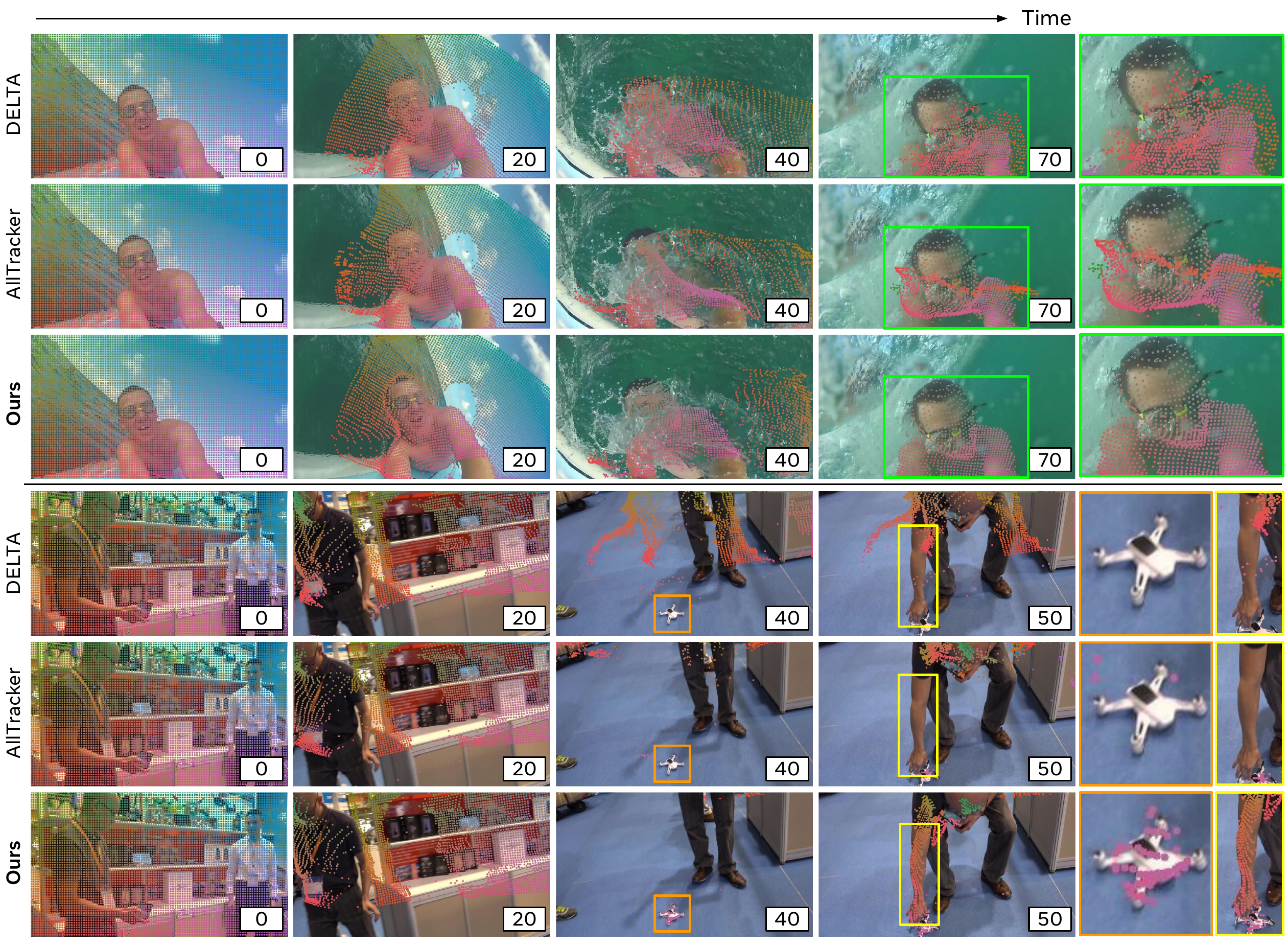}
\caption{\textbf{Qualitative comparison on challenging sequences involving non-rigid water motion and small object tracking.} 
Columns compare {DELTA}, {AllTracker}, and {Ours}. 
The top sequence exhibits large non-rigid motion and drastic appearance changes as the person dives into the water. 
The bottom sequence features large camera motion where objects repeatedly go out of frame, requiring the tracking of a small object. 
Our method maintains a consistent track in these difficult conditions, whereas {DELTA} and {AllTracker} fail to recover or exhibit drift. 
Notably, our approach successfully tracks the small object in the bottom sequence, thanks to our high-resolution feature maps that preserve fine details. Numbers in lower-right boxes indicate frame numbers.}
\label{fig:supp_track_compare}
\end{figure*}

%% file: supp_figs/trackothers.tex
\begin{figure*}[t]
\centering
\includegraphics[width=0.99\linewidth]{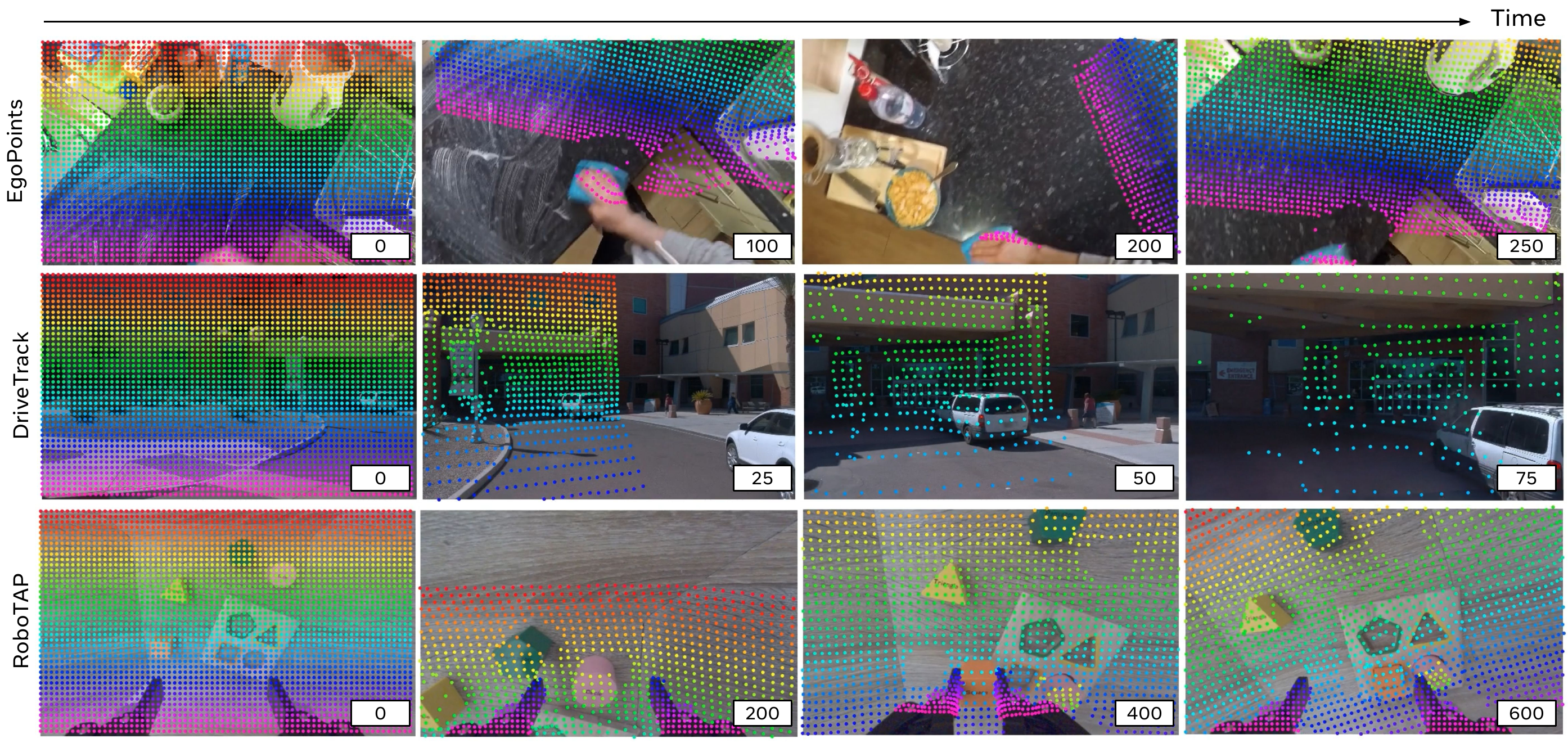}
\caption{\textbf{Generalization to diverse video domains.} Our model demonstrates high tracking accuracy across a variety of domains, including ego-centric videos (top), self-driving scenes (middle), and robotics manipulation tasks (bottom). Notably, our approach is robust over long temporal extents, successfully maintaining tracks in sequences as long as 600 frames (see bottom row).}
\label{fig:supp_track_generalize}
\end{figure*}

%% file: supp_figs/flow_results_supp.tex
\begin{figure*}[t]
\centering
\includegraphics[width=0.99\linewidth]{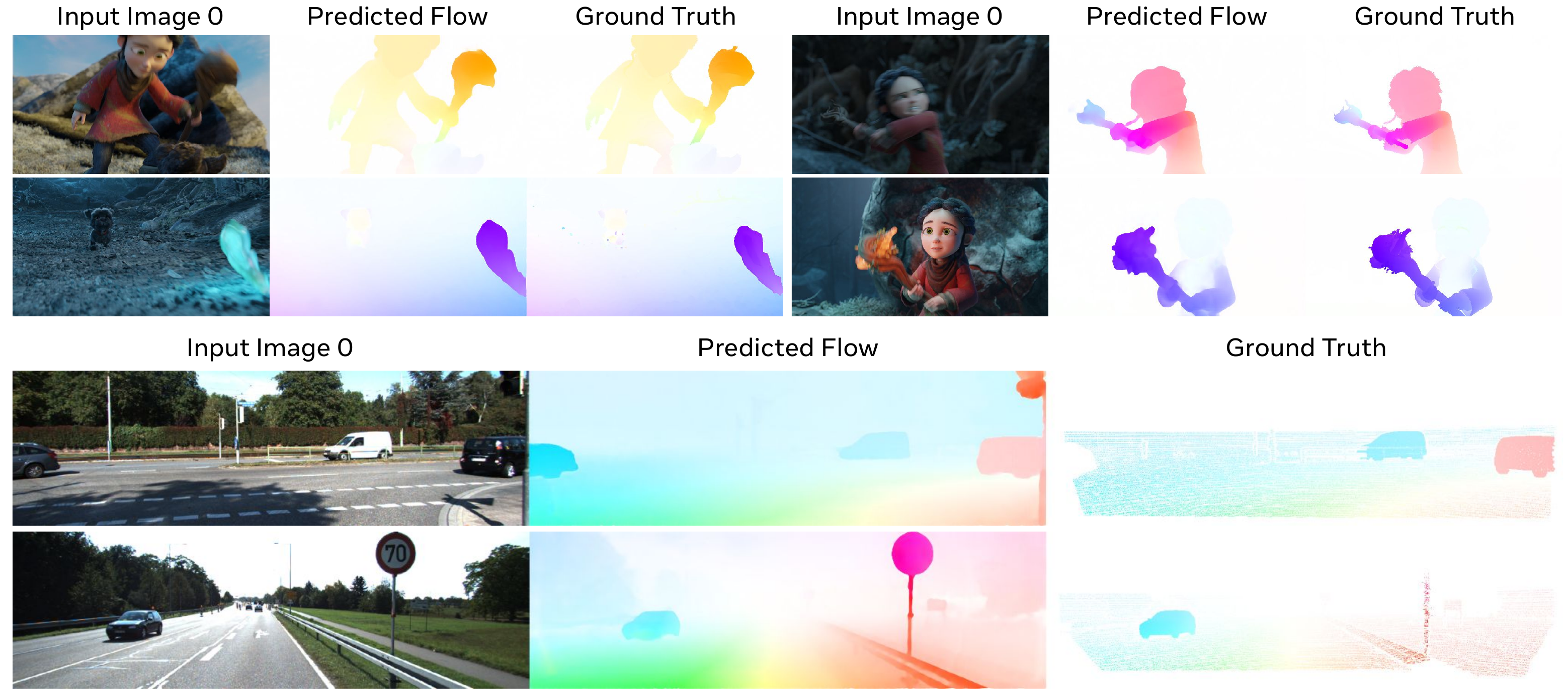}
\caption{\textbf{Additional qualitative optical flow results.} 
Complementing the Sintel results in the main paper, we visualize performance on the Spring dataset (top two rows) and the KITTI dataset (bottom two rows). 
Our model predicts high-quality optical flow with sharp motion boundaries and accurate dense correspondence that closely matches the ground truth. 
It is important to note that we employ the exact same tracking model used throughout the paper; it was \textit{not} trained on any optical flow datasets, including Spring or KITTI, demonstrating its strong generalization capability.}
\vspace{-1em}
\label{fig:supp_flow}
\end{figure*}